\documentclass{article}

\usepackage{arxiv}

\usepackage[utf8]{inputenc} 
\usepackage[T1]{fontenc}    
\usepackage{hyperref}       
\usepackage{url}            
\usepackage{booktabs}       
\usepackage{amsfonts}       
\usepackage{nicefrac}       
\usepackage{microtype}      
\usepackage{graphicx}
\usepackage[square,sort,comma,numbers]{natbib}
\usepackage{doi}
\usepackage{subcaption}
\usepackage{amssymb, amsmath, amsthm}
\usepackage{algorithm}
\usepackage{algpseudocode}
\usepackage{appendix}
\usepackage{enumitem}
\usepackage{graphbox}
\usepackage[english]{babel}
\usepackage{amsthm}
\newtheorem{theorem}{Theorem}
\newtheorem{definition}{Definition}
\newtheorem{prop}[theorem]{Proposition}

\title{Paramixer: Parameterizing Mixing Links in Sparse Factors Works Better than Dot-Product Self-Attention}

\date{} 					

\newcommand*\samethanks[1][\value{footnote}]{\footnotemark[#1]}
\author{ Tong Yu\thanks{Equal contribution}, \  Ruslan Khalitov\samethanks, \  Lei Cheng, \ Zhirong Yang\thanks{Corresponding author, \texttt{zhirong.yang@ntnu.no}}\\
Norwegian University of Science and Technology}

\renewcommand{\shorttitle}{Parameterizing Mixing Links in Sparse Factors Works Better than Dot-Product Self-Attention}

\hypersetup{
	pdftitle={Paramixer: Parameterizing Mixing Links in Sparse Factors Works Better than Dot-Product Self-Attention},
	pdfsubject={},
	pdfauthor={Tong Yu, Ruslan Khalitov, Lei Cheng, and Zhirong Yang},
	pdfkeywords={parameterization, mixing links, sparse matrix, matrix factorization},
}

\begin{document}
	\maketitle

\newcommand{\matA}{\mathbf{A}}
\newcommand{\matB}{\mathbf{B}}
\newcommand{\matC}{\mathbf{C}}
\newcommand{\matD}{\mathbf{D}}
\newcommand{\matE}{\mathbf{E}}
\newcommand{\matF}{\mathbf{F}}
\newcommand{\matG}{\mathbf{G}}
\newcommand{\matH}{\mathbf{H}}
\newcommand{\matI}{\mathbf{I}}
\newcommand{\matK}{\mathbf{K}}
\newcommand{\matL}{\mathbf{L}}
\newcommand{\matM}{\mathbf{M}}
\newcommand{\matN}{\mathbf{N}}
\newcommand{\matO}{\mathbf{O}}
\newcommand{\matP}{\mathbf{P}}
\newcommand{\matQ}{\mathbf{Q}}
\newcommand{\matR}{\mathbf{R}}
\newcommand{\matS}{\mathbf{S}}
\newcommand{\matT}{\mathbf{T}}
\newcommand{\matU}{\mathbf{U}}
\newcommand{\matV}{\mathbf{V}}
\newcommand{\matW}{\mathbf{W}}
\newcommand{\matX}{\mathbf{X}}
\newcommand{\matY}{\mathbf{Y}}
\newcommand{\matZ}{\mathbf{Z}}
\newcommand{\matg}{\mathbf{g}}

\newcommand{\calA}{\mathcal{A}}
\newcommand{\calB}{\mathcal{B}}
\newcommand{\calC}{\mathcal{C}}
\newcommand{\calD}{\mathcal{D}}
\newcommand{\calE}{\mathcal{E}}
\newcommand{\calF}{\mathcal{F}}
\newcommand{\calG}{\mathcal{G}}
\newcommand{\calH}{\mathcal{H}}
\newcommand{\calI}{\mathcal{I}}
\newcommand{\calJ}{\mathcal{J}}
\newcommand{\calK}{\mathcal{K}}
\newcommand{\calL}{\mathcal{L}}
\newcommand{\calM}{\mathcal{M}}
\newcommand{\calN}{\mathcal{N}}
\newcommand{\calO}{\mathcal{O}}
\newcommand{\calP}{\mathcal{P}}
\newcommand{\calQ}{\mathcal{Q}}
\newcommand{\calR}{\mathcal{R}}
\newcommand{\calS}{\mathcal{S}}
\newcommand{\calT}{\mathcal{T}}
\newcommand{\calU}{\mathcal{U}}
\newcommand{\calV}{\mathcal{V}}
\newcommand{\calW}{\mathcal{W}}
\newcommand{\calX}{\mathcal{X}}
\newcommand{\calY}{\mathcal{Y}}
\newcommand{\calZ}{\mathcal{Z}}

\newcommand{\bbA}{\mathbb{A}}
\newcommand{\bbB}{\mathbb{B}}
\newcommand{\bbR}{\mathbb{R}}
\newcommand{\bbZ}{\mathbb{Z}}
\newcommand{\bbE}{\mathbb{E}}
\newcommand{\bbH}{\mathbb{H}}

\newcommand{\veca}{\mathbf{a}}
\newcommand{\vecb}{\mathbf{b}}
\newcommand{\vecc}{\mathbf{c}}
\newcommand{\vecd}{\mathbf{d}}
\newcommand{\vece}{\mathbf{e}}
\newcommand{\vecf}{\mathbf{f}}
\newcommand{\vecg}{\mathbf{g}}
\newcommand{\vech}{\mathbf{h}}
\newcommand{\veci}{\mathbf{i}}
\newcommand{\vecj}{\mathbf{j}}
\newcommand{\veck}{\mathbf{k}}
\newcommand{\vecl}{\mathbf{l}}
\newcommand{\vecm}{\mathbf{m}}
\newcommand{\vecn}{\mathbf{n}}
\newcommand{\veco}{\mathbf{o}}
\newcommand{\vecp}{\mathbf{p}}
\newcommand{\vecq}{\mathbf{q}}
\newcommand{\vecr}{\mathbf{r}}
\newcommand{\vecs}{\mathbf{s}}
\newcommand{\vect}{\mathbf{t}}
\newcommand{\vecu}{\mathbf{u}}
\newcommand{\vecv}{\mathbf{v}}
\newcommand{\vecw}{\mathbf{w}}
\newcommand{\vecx}{\mathbf{x}}
\newcommand{\vecy}{\mathbf{y}}
\newcommand{\vecz}{\mathbf{z}}

\newcommand{\vecalpha}{\boldsymbol{\alpha}}
\newcommand{\vecbeta}{\boldsymbol{\beta}}
\newcommand{\veceta}{\boldsymbol{\eta}}
\newcommand{\vectheta}{\boldsymbol{\theta}}
\newcommand{\vecphi}{\boldsymbol{\phi}}
\newcommand{\vecpsi}{\boldsymbol{\psi}}
\newcommand{\vecrho}{\boldsymbol{\rho}}
\newcommand{\vectau}{\boldsymbol{\tau}}
\newcommand{\vecmu}{\boldsymbol{\mu}}
\newcommand{\veceps}{\boldsymbol{\epsilon}}
\newcommand{\vecxi}{\boldsymbol{\xi}}
\newcommand{\vecPhi}{\boldsymbol{\Phi}}
\newcommand{\vecDelta}{\boldsymbol{\Delta}}

\newcommand{\matDelta}{\boldsymbol{\Delta}}
\newcommand{\matEta}{\boldsymbol{\eta}}
\newcommand{\matOmega}{\boldsymbol{\Omega}}
\newcommand{\matPhi}{\boldsymbol{\Phi}}
\newcommand{\matPsi}{\boldsymbol{\Psi}}
\newcommand{\matTheta}{\boldsymbol{\Theta}}
\newcommand{\matLambda}{\boldsymbol{\Lambda}}
\newcommand{\matSigma}{\boldsymbol{\Sigma}}
\newcommand{\matzero}{\mathbf{0}}
\newcommand{\IndexSetI}{\mathcal{I}}
\newcommand{\grad}{\mathcal{\nabla}}

\newcommand{\vecone}{\mathbf{1}}
\newcommand{\veczero}{\mathbf{0}}

\def\maximize{\mathop{{\mathgroup\symoperators maximize}}}
\def\Maximize{\mathop{{\mathgroup\symoperators Maximize}}}
\def\minimize{\mathop{{\mathgroup\symoperators minimize}}}

\def\approach{\mathop{{\mathgroup\symoperators \longrightarrow}}}
\def\defineoperator{\mathop{{\mathgroup\symoperators =}}}
\newcommand{\define}{\defineoperator^{\text{def}}}

\newcommand{\Tr}{\text{Tr}}
\newcommand{\trace}{\text{trace}}
\newcommand{\diag}{\text{diag}}
\newcommand{\gradWJ}{\nabla_{\scriptscriptstyle{\matW}}\calJ}
\newcommand{\const}{\text{constant}}
\newcommand{\fracpartial}[2]{\frac{\partial #1}{\partial  #2}}

\newcommand{\defeq}{\stackrel{\text{def}}{=}}

\newcommand{\Xh}{\widehat{X}}

\begin{abstract}
Self-Attention is a widely used building block in neural modeling to mix long-range data elements. Most self-attention neural networks employ pairwise dot-products to specify the attention coefficients. However, these methods require $O(N^2)$ computing cost for sequence length $N$. Even though some approximation methods have been introduced to relieve the quadratic cost, the performance of the dot-product approach is still bottlenecked by the low-rank constraint in the attention matrix factorization. In this paper, we propose a novel scalable and effective mixing building block called Paramixer. Our method factorizes the interaction matrix into several sparse matrices, where we parameterize the non-zero entries by MLPs with the data elements as input. The overall computing cost of the new building block is as low as $O(N \log N)$. Moreover, all factorizing matrices in Paramixer are full-rank, so it does not suffer from the low-rank bottleneck. We have tested the new method on both synthetic and various real-world long sequential data sets and compared it with several state-of-the-art attention networks. The experimental results show that Paramixer has better performance in most learning tasks.\footnote{https://github.com/wiedersehne/Paramixer}
\end{abstract}


\section{Introduction}
\label{sec:intro}

Transformer models have been widely used on many tasks such as text classification \cite{vaswani2017attention}, text summarization, promoter region prediction \cite{zaheer2020big}, and image classification \cite{dosovitskiy2020vit}. The main engine in Transformer is the self-attention mechanism, which can work in parallel to mix long-range tokens in a long sequence. This fundamental innovation eliminated the sequential dependency in recurrent neural networks and was used as a building block for many powerful models, such as Bert \cite{devlin2018bert}, GPT\cite{brown2020gpt} and Ernie\cite{sun2019ernie}. 

However, the original self-attention is not scalable because it requires computing and storing all pairwise dot-products, which incurs $O(N^2)$ cost for sequence length $N$. The scalability issue significantly restricted the application of neural models based on self-attention.

Various methods have been introduced to alleviate the quadratic cost of full attention. Some of them attempt to shorten the sequence length \cite{linformer, enformer}, even though much information is lost. Others try to break up the softmax by a certain kernel factorization.  Another family of methods sparsify the attention matrix with predefined attention \cite{beltagy2020longformer,zaheer2020big,ainslie2020etc,child2019generating,tay2019lightweight}. However, most Transformer variants stick to the dot-product self-attention, of which the expressive power is restricted by the low-rank bottleneck \cite{mhabottleneck} because the dimensionality of the dot-product space is much smaller than the sequence length. Therefore, they cannot accurately model the transformation if the attention is intrinsically high-rank.


This paper proposes a scalable and effective attention building block called Paramixer without dot-product and softmax. Our method directly parameterizes the mixing links in several sparse factors to form an attention matrix, where all factorizing matrices are full-rank. Therefore Paramixer does not suffer from the low-rank bottleneck. We present two ways to specify the non-zero positions in each sparse factor. Both lead to an economical approximation of the full attention matrix, with the computing cost as low as $O(N\log N)$. As a result, our method can easily model very long sequential data.

We have tested Paramixer on various sequence data sets and compared it with many popular self-attention neural networks based on dot-products. The experimental results show that Paramixer gets the best performance on very long sequence tasks, including synthetic data inference, Genome classification, and character-level long document classification. Paramixer also achieves state-of-art accuracy on the public Long Range Arena benchmark tasks.

We organize the rest of the paper as follows. Section 2 investigates dot-product self-attention and its related work. Section 3 introduces the development clue and model architecture of Paramixer. The experimental settings and results are presented in Section 4, and we conclude the paper in Section 5.

\section{Related Work}
\label{sec:relwork}

Self-attention (SA) is a building block in neural networks which enables long range interaction between elements in a sequence. The most widely used SA architecture is called Transformer \cite{vaswani2017attention}, where the self-attention matrix is constructed using scaled dot-product followed by softmax. Given an input sequence of $N$ elements encoded in $X\in\mathbb{R}^{N\times {d}}$, a self-attention building block calculates a weighted average of feature representations $V\in\mathbb{R}^{N\times {d_v}}$, where the weights are the result of scaled dot-product of $Q\in\mathbb{R}^{N\times {D}}$ and $K\in\mathbb{R}^{N\times {D}}$: $Q = XW_q$, $K = XW_k$, and $V = XW_v$.
Then the self-attention in Transformer is
\begin{align}
\label{eq:self-attention}
    \text{Self-Attention}(Q,K,V) = AV,
\end{align}
where 
\begin{align}
\label{eq:scaleddotproduct}
A = \text{softmax}\left(\frac{QK^T}{\sqrt{D}}\right)
\end{align}
and the softmax applies row-wise on the scaled dot-products.

The attention matrix in Eq.~\ref{eq:scaleddotproduct} is not scalable because it requires computing and storing $N^2$ attention values, which is infeasible for a large $N$. The quadratic cost becomes a significant bottleneck when applying self-attention applications for long sequences. Many research teams have proposed Transformer variants to relieve this problem.

The first branch of methods attempts to reduce $N$.
The Linformer approximation \cite{wang2020linformer} uses random projection to reduce the rows of $K$ and $V$ from $N$ to $r$ with $r<N$. However, because $r\propto\epsilon^{-2}$ with $\epsilon$ the approximation error bound, Linformer has to use a large $r$ to achieve satisfactory approximation quality.
Another method called Enformer \cite{enformer} shortens the sequence length by convolutional network pooling.

The second branch of methods tries to break up the softmax by a certain kernel factorization $A\approx\phi(Q)\phi(K)^T$, where $\phi_Q$ and $\phi_K\in\bbR^{N\times r'}$ with $r'<N$. Then by the association rule of multiplication, we can calculate $\phi(Q)\left[\phi(K)^TV\right]$ and avoid the quadratic cost. For example, Nystr\"omformer \cite{nystromformer} uses a few landmarks as surrogates to construct $\phi(Q)$ and $\phi(K)$. Linear Transformer \cite{katharopoulos2020transformers} directly chooses $\phi(x)=\text{elu}(x)+1$. Performer \cite{performer} uses $r'$ orthogonal random features to obtain $\phi(Q)$ and $\phi(K)$. Random features were also used in another work \cite{peng2021random}, with gating mechanism combined.

The third branch of methods chooses only a subset of $(i,j)$ pairs in each attention layer. These include Sparse Transformers \cite{child2019generating,ainslie2020etc} that use a set of neighboring tokens, Sinkhorn Transformer \cite{tay2019lightweight} that uses blockwise sparsity, Longformer \cite{beltagy2020longformer} that uses dilated sparse connections, and BigBird \cite{zaheer2020big} that uses both blockwise and dilated sparse connections.

\begin{figure*}[t]
\newcommand{\sqimgwidth}{4.0cm}
\newcommand{\parawidth}{4.8cm}
	\begin{center}
	\begin{tabular}{cccc}
		\includegraphics[width=\sqimgwidth]{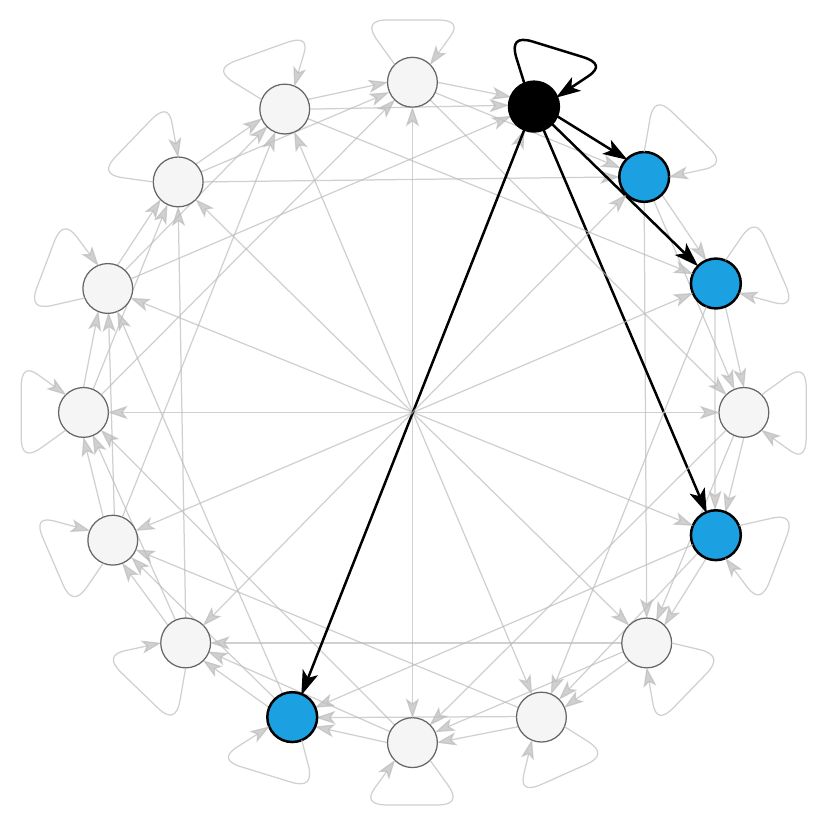} &
		\includegraphics[width=\sqimgwidth]{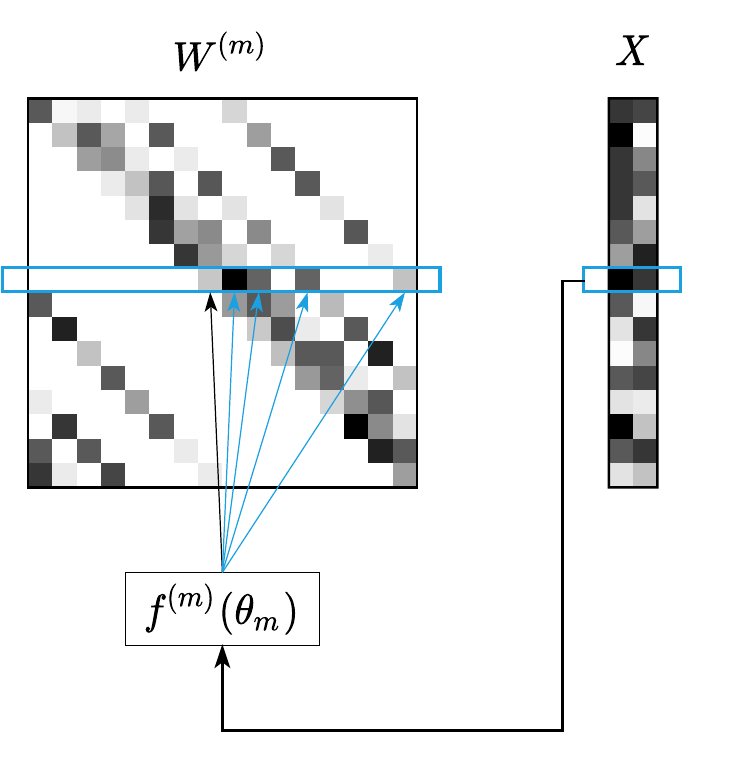} &
		\includegraphics[width=\sqimgwidth]{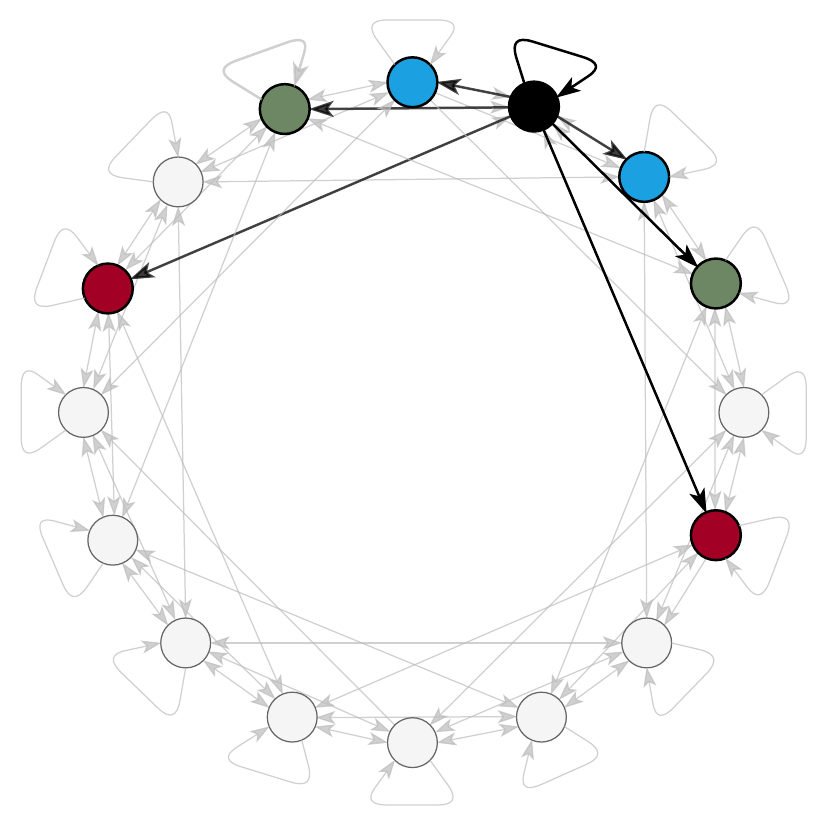} &
		\includegraphics[width=\sqimgwidth]{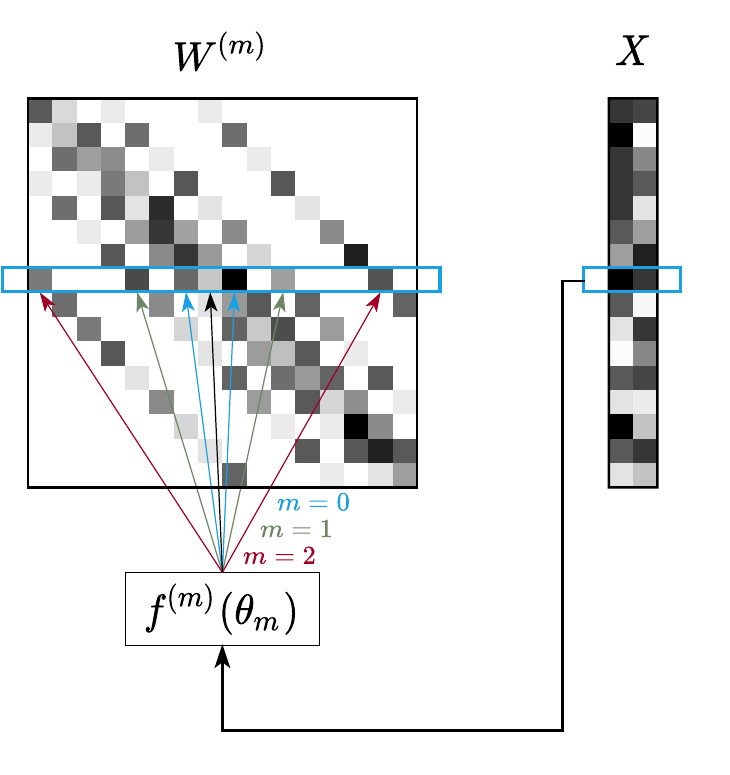} \\
		(a) & (b) & (c) & (d)\\
	\end{tabular}
	\end{center}
	\caption{Illustration of (a \& b) the CHORD and (c \& d) the CDIL protocols for $N=16$. Each node in the circular graph represents a sequence element. The links between nodes correspond to the non-zero entries in $W^{(m)}$ (here $m=1$) output from $f^{(m)}$. Note that the sparse structure of all factors in CHORD is the same, while it varies at different $m$'s in CDIL.}
	\label{fig:paramixer_factor}
\end{figure*}

\section{Paramixer}

Despite many variants of Transformers, there are still several drawbacks because they stick to dot-product + softmax or their approximation. Below we discuss the disadvantages of these two components and propose our solution without them.

\subsection{Drawbacks of dot-products and softmax}
The first drawback is the low-rank bottleneck: dot-product self-attention requires $D\ll N$; otherwise, the $Q$ and $K$ matrices are unaffordable for a large $N$. Bhojanapalli et al.~have shown that the low-rank bottleneck restricts the representation power and attention performance \cite{mhabottleneck}. However, their workaround that uses $D=N$ does not apply for long sequences due to the quadratic cost.

Another limitation comes from the pairwise definition of dot-product. Although dot-products have a theoretical connection to Reproducing Kernel Hilbert Space (RKHS), self-attention cannot benefit from more than two factors in the matrix product because the kernels are defined in pairs.

Another key component, softmax, in Transformer is also problematic. First, such nonlinearity over dot-products leads to quadratic cost. Comprehensive approximation methods are required to approximate softmax for more economical matrix products.

Second, softmax limits the mixing capability. The softmax layer output probabilities and the mixing result are thus constrained in the convex hull of the existing elements. The constraint is more severe in sparse attention methods such as \cite{child2019generating,ainslie2020etc,tay2019lightweight,beltagy2020longformer,zaheer2020big}, where the mixing result is in a rather constrained convex hull of a few involved elements.

Moreover, softmax is incompatible with sparse attention. The latter was introduced to relieve the quadratic computing cost caused by softmax. However, after each sparse attention layer with softmax, the network can output probabilities for only a few sequence elements.

\subsection{Parameterizing mixing links}

Seeing the drawbacks of dot-products and softmax, we rethink self-attention as a transformation block and redesign the neural model. First, we drop the softmax because it is not compulsory for the mixing function but limits the mixing capability. Then the transformation becomes $V^\text{new}=A V$ for an unconstrained mixing matrix $A$.

Next, we consider parameterizing the mixing links or coefficients for each matrix row $A_{i:}=f(X_i; \theta)$, where $f: \bbR^d\mapsto\bbR^N$ is a neural network with weights $\theta$. However, such simple parameterization does not solve the quadratic cost. We thus consider a sparse factorization of the mixing matrix $A$:
\begin{align}
    \label{eq:fac}
    A=\prod_{m=1}^M W^{(m)},
\end{align}
where each sparse factor $W^{(m)}$ is a full-rank sparse square matrix. Then we parameterize each sparse factor $W^{(m)}$ as
\begin{align}
    W^{(m)}_{i:} = f^{(m)}(X_i; \theta_m),
\end{align}
where $i=1,\dots,N$ and $f:\bbR^d\mapsto\bbR^K$ is a Multilayer Perceptron (MLP) that outputs the $K$ non-zero entries\footnote{Non-zero entries are those stored, including both non-zeros and explicit zeros.} of each row in $W^{(m)}$.
If the total number of non-zeros in the factors is much smaller than $N^2$, we obtain a new economical self-attention method without dot-product and softmax. We call the new method Paramixer.

There are different ways or protocols to specify the sparse structure or the non-zero entries. We have studied two protocols and present them below. For short, we abbreviate $f^{(m)}_{ik}\defeq\left[f^{(m)}(X_i; \theta_m)\right]_k$.

The first protocol CHORD modifies from an algorithm with the same name in peer-to-peer lookup service \cite{chord}, where the $i$-th row of each sparse factor $W^{(m)}$ is parameterized as
\begin{align}
\label{eq:CHORD}
    W^{(m)}_{ij}=
    \begin{cases}
    f^{(m)}_{i1} & \text{if }j=i \\
    f^{(m)}_{ik} & \text{if }j=i+2^{k-2}\mod N \\
    0 & \text{otherwise}.
    \end{cases}
\end{align}
where $j=1,\dots,N$ and $k = 1,...,K$. That is, each row has $K$ non-zeros, and in total all $W^{(m)}$'s have $MNK$ non-zeros. In this work, we follow the original CHORD algorithm to use $K=M=\log N$. Therefore the number of stored entries is $N\log^2N$. 

Each factor $W^{(m)}$ can be treated as the adjacency matrix of a directed graph, where the $(i,j)$-th non-zero entry can be seen as a link from $i$ to $j$. The graph visualization and the parameterization are illustrated in Figure \ref{fig:paramixer_factor}.

The product of the factorizing matrices corresponds to the connections in the circular graph after multiple sparse factors. Theorem 2 in \cite{chord} guarantees that the graph will become complete with high probability after $M=\log N$ sparse factors. That is, the resulting $A$ matrix will become full after the matrix product.

\begin{figure*}[t]
	\begin{center}
		\includegraphics[width=\textwidth]{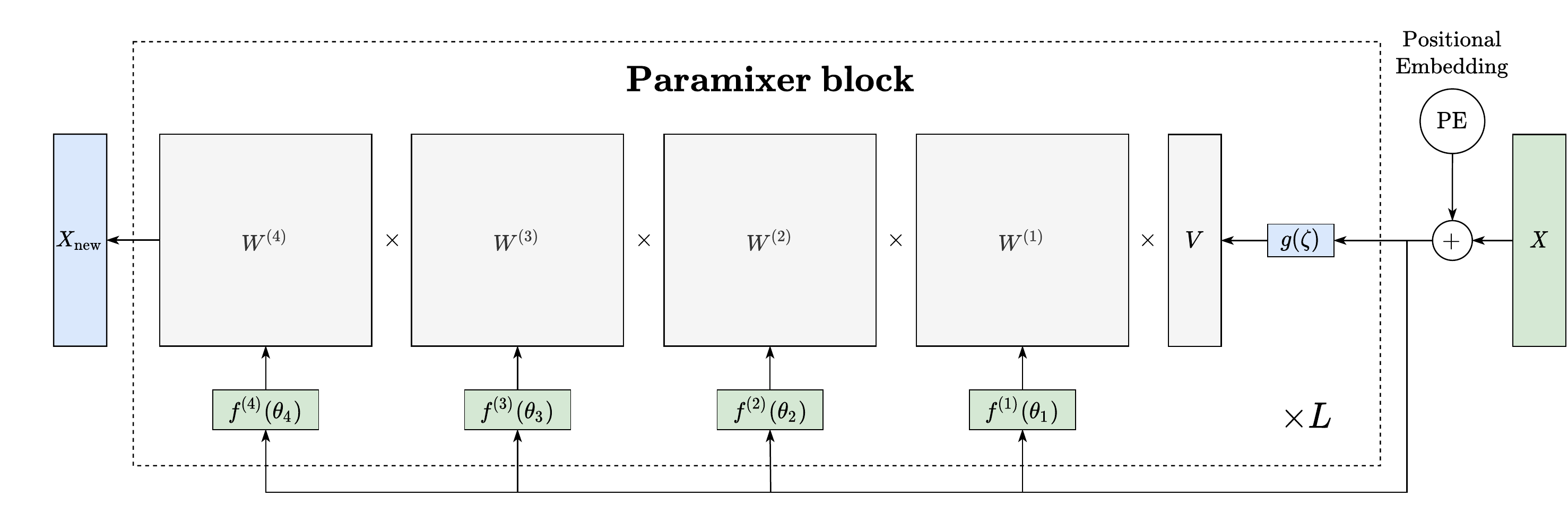}
	\end{center}
	\caption{Illustration of an example Paramixer neural network ($M=4$). After adding the positional embedding, it applies $L$ Paramixer blocks and obtains the transformed tensor $X_\text{new}$.}
	\label{fig:paramixer_nn}
\end{figure*}

The second protocol CDIL (Circular DILated) originates from Temporal Convolution Networks (TCN) \cite{tcn}, where we modify the dilated connections at both sides and along a circular graph to ensure the receptive fields are symmetric. The $i$-th row of $W^{(m)}$ is parameterized as 
\begin{align}
\label{eq:CDIL}
    W^{(m)}_{ij}=
    \begin{cases}
    f^{(m)}_{i1} & \text{if }j=i \\
    f^{(m)}_{ik} & \text{if }j=i+p_{k-1}2^m\mod N\\
    0 & \text{otherwise},
    \end{cases}
\end{align}
where $p=\left[1,2,...,\frac{K-1}{2},-1,-2,-\frac{K-1}{2}\right]$ and $k=2,\dots,K$.
Similar to CHORD, the CDIL protocol includes self links but the other links appear on both sides, and the dilation $2^m$ varies at different $m$'s. The CDIL protocol and parameterization are illustrated in Figure \ref{fig:paramixer_factor}.

In CDIL, each factor contains $KN$ non-zero entries, and in total, there are $KN\log N$ non-zeros if $M=\log N$, which is more economical than CHORD if $K<\log N$. It can be proven that the product $A=\prod_{m=1}^MW^ {(m)}$ is a full matrix with the CDIL protocol. 

The following proposition states that the factorizing matrices constructed by the two protocols are full-rank. The proof in given the supplemental document.
\begin{prop}
$W^{(1)},\dots,W^{(M)}$ constructed in Eq.~\ref{eq:CHORD} or Eq.~\ref{eq:CDIL} are in general full-rank.
\end{prop}

\subsection{Paramixer neural networks}
\label{sec:paramixer_nn}
A Paramixer block transforms an $N\times d$ tensor $X$ to another tensor of the same size. See Figure \ref{fig:paramixer_nn}. Here we use an MLP $g:\bbR^d\mapsto\bbR^d$ with weights $\zeta$ to mix the columns of the input tensor, which is similar to the product with $W_v$ in Transformer. In this work, we used simple three-layer MLPs (Linear-GELU-Linear) for both $f^{(m)}$'s and $g$. Stacking $L$ such blocks forms a neural network backbone, denoted by ParamixerNet, for representation learning. 

We have tried two options for parameterization in multi-block setting: $W^{(l,m)}=f^{(l,m)}\left(X^{(l-1)}; \theta^{(l)}_m\right)$ and $W^{(l,m)}=f^{(l,m)}\left(X^{(0)}; \theta^{(l)}_m\right)$, where the superscript $l$ indexes the $l$-th block and $X^{(l-1)}$ is the input to the $l$-th block, with $X^{(0)}=X+\text{PE}$. We find that the latter option makes the network easier to train and works better.

The ParamixerNet output can then be fed to a loss function $\calJ$, and overall, the learning task can be formulated as the following optimization problem:
\begin{align}
\label{eq:paramixernet}
    \minimize_{\Theta}~~\calJ(\text{ParamixerNet}(X+\text{PE}; \Theta)),
\end{align}
where $\Theta=\left\{\theta^{(l)}_1,\dots,\theta^{(l)}_M, \zeta^{(l)}\right\}_{l=1}^{L}$.
The optimization can be implemented with back-propagation, and a gradient-based algorithm such as Adam \cite{adam}.

\section{Experiments}
\label{sec:exps}
We conducted four groups of experiments. In the first group, we demonstrate the scalability of Paramixer on long synthetic sequences with lengths up to tens of thousands of positions. Then, we tested the performance of Paramixer on the pubic Long Range Arena benchmark data sets. In the third group, we built a character-level document classification task to evaluate if Paramixer can handle real-world long text sequences with tens of thousands of tokens on average. Finally, we showcase if Paramixer performs well in modeling long genome sequences. We ran all experiments on a Linux machine with 3$\times$NVIDIA Tesla V100 32GB, Intel Xeon Gold 6240 CPU @ 2.60GHz processors, with 754GB of system memory. 

\subsection{Synthetic Scalability Test}
In this section, we examine the scalability of Paramixer and compare its performance with several competitors. We used two synthetic data sets composed of long sequences for supervised learning tasks. An experimental setup was inspired by  \cite{hochreiter1997long}, where similar synthetic sequences appeared for scalability tests. The details of both tasks are given below:
\begin{itemize}
    \item \textit{Adding Problem}. This is a sequence regression task. Each element of an input sequence is a pair of numbers $(a_i, b_i)$, where $a_i\sim U(-1, 1)$, $b_i\in \{0, 1\}$, $i=1,\dots, N$. We generated signals at two randomly selected positions $t_1$ and $t_2$ such that $b_{t_1}=b_{t_2}=1$ and $b_i = 0$ elsewhere. The learning target is $y = 0.5+\dfrac{a_{t_1} + a_{t_2}}{4}$. For example, an input sequence $[(0.5, 1), (-0.2, 0), (0.2,1), (-0.8,0), (0.6,1)]$ will have the learning target $y=0.825$. Unlike \cite{hochreiter1997long}, we did not restrict the $t_1$ and $t_2$ choice and made the task more challenging. That is, the relevant signals can appear either locally or at a great distance from each other. In evaluation, a network prediction $\hat{y}$ is considered correct if $|y - \hat{y}| < 0.04$.
    \item \textit{Temporal Order}. This is a sequence classification task. Each sequence consists of randomly chosen symbols from the alphabet $\{a, b, c, d, X, Y\}$, where the first four are noise symbols. Each sequence has two signal symbols, either $X$ or $Y$, which appear at two arbitrary positions. The four target classes correspond to the ordered combinations of the signal symbols $(X,X)$, $(X, Y)$, $(Y,X)$, and $(Y, Y)$. For example, an input sequence $[a,d,Y,c,b,a,Y,c,d]$ should be classified as Class 3.
\end{itemize}
We generated data of different sequence lengths for each problem: from $N=128$ to $N=2^{15}$, progressively increasing the length by the factor of two. For each sequence length, a model can access 100\,000 training sequences and 5\,000 testing instances for evaluation. 

We compared Paramixer with a group of popular methods based on scaled dot-product attention (referred as X-formers), including Linformer \cite{linformer}, Performer \cite{performer}, Reformer \cite{kitaev2020reformer} and Nystr\"omformer, which all \cite{nystromformer} have claimed to be scalable. For completeness, we also included the original Transformer \cite{vaswani2017attention}. We used the open-source PyTorch implementations\footnote{available at \url{https://github.com/lucidrains}} of these models.

We fine-tuned the main hyperparameters in a standard cross-validation manner for Paramixer and X-formers, including the number of layers and heads, dimensionality of the token embedding, and query/key/value dimensions. For the Temporal Order problem, we directly fed the data instances to the embedding layers. For the Adding problem, the input data was only two-dimensional, and one of them was real-valued. Directly using such a low-dimensional embedding space would limit the expressive power. So we added a linear layer to augment the dimensionality to allow sufficient freedom for the scaled dot-products in the X-former architectures. All the models were optimized using the Adam optimizer \cite{adam} with the learning rate of 0.001 using a batch size of 40.

\setlength\tabcolsep{1.5pt}
\newcommand{\sqimgwidth}{7.5cm}
\begin{figure}[t]
    \centering
    \includegraphics[width=\sqimgwidth]{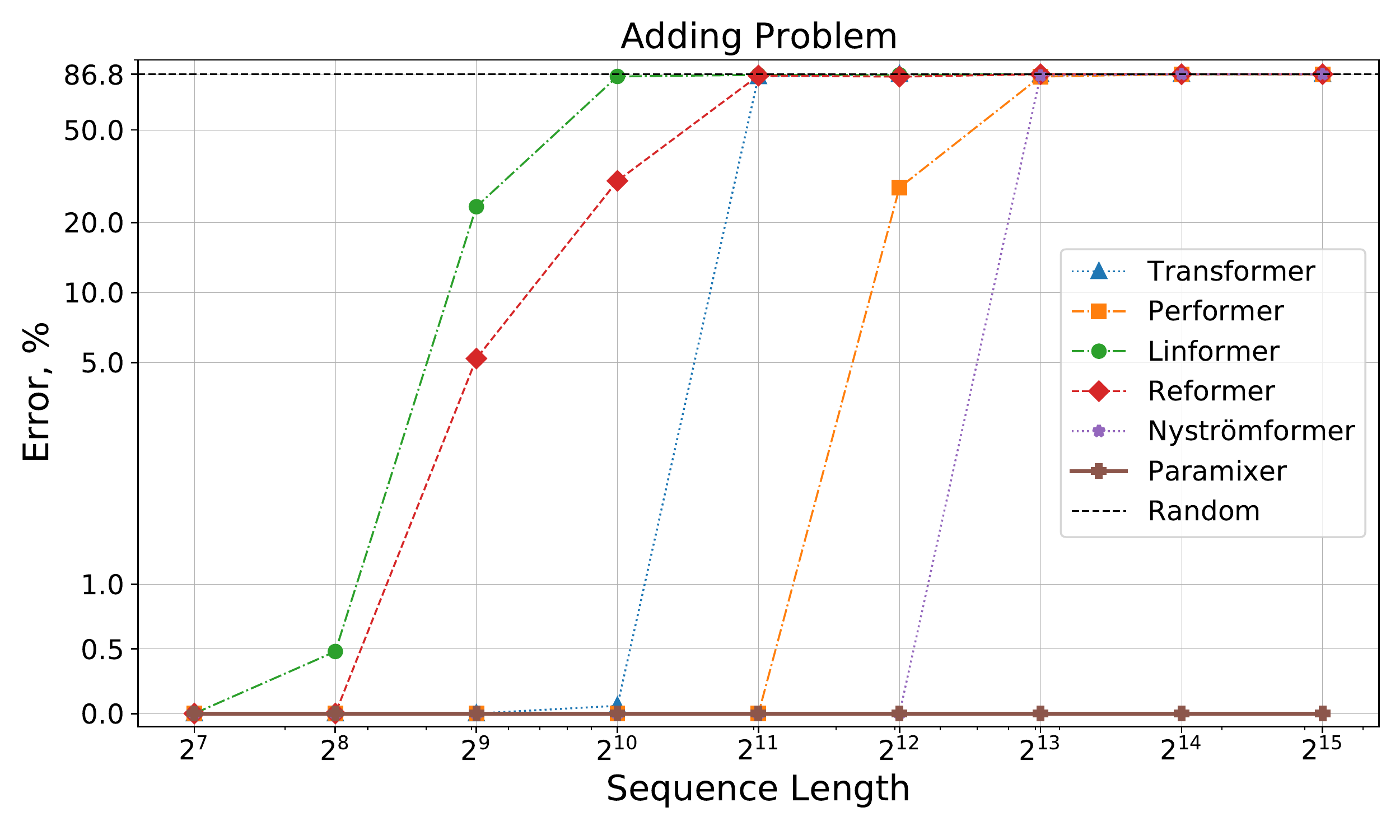}
	\includegraphics[width=\sqimgwidth]{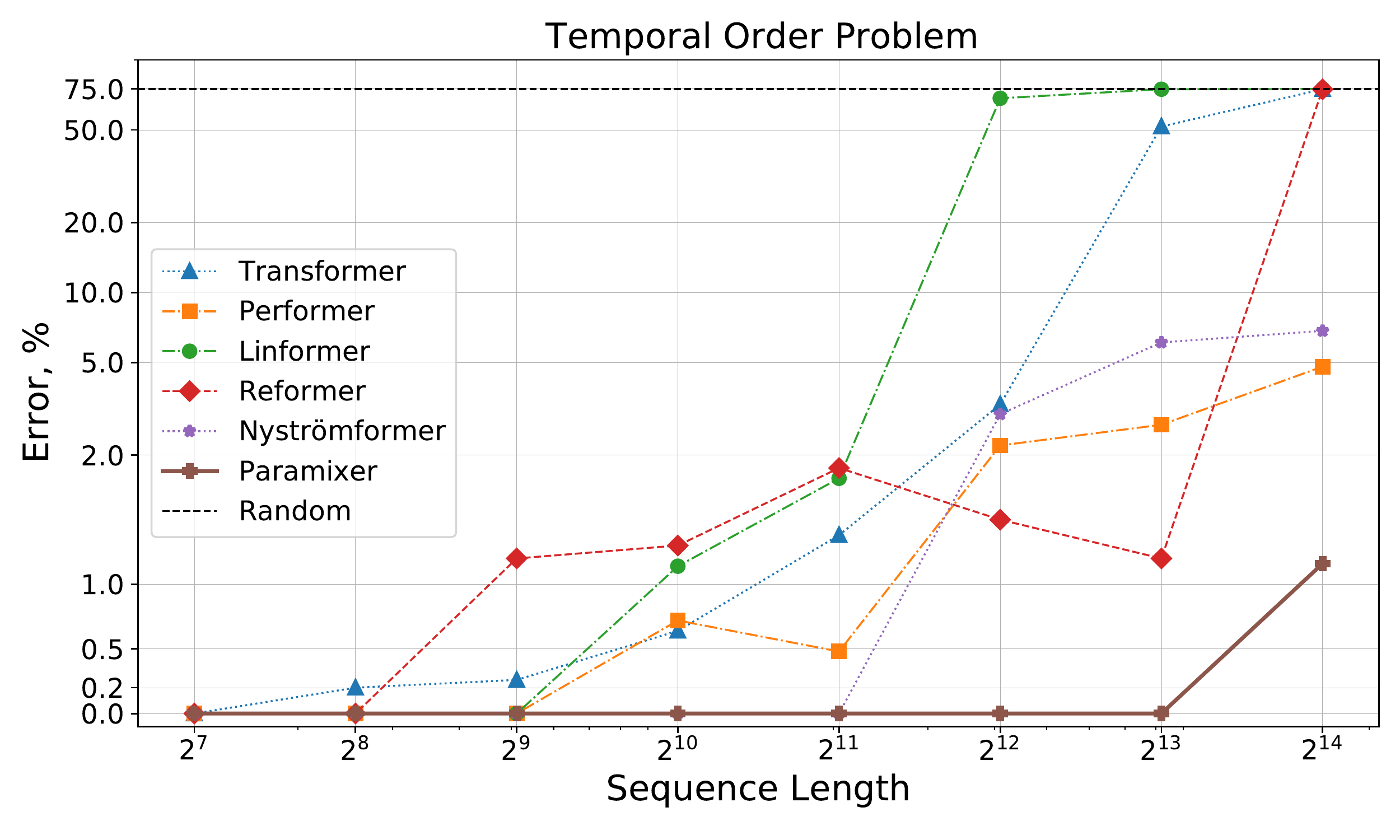}\\
    \caption{Error percentage of Paramixer and the X-formers for both the Adding problem and the Temporal Order problem with increasing sequence lengths.}
    \label{fig:acc_synth}
\end{figure}

The results are shown in Figure \ref{fig:acc_synth}. For the Adding problem, we see that all models obtain 100\% accuracy for $N\leq256$. The X-former models become weak or ineffective when scaling to longer sequences, while Paramixer still performs well. Linformer and Reformer get an error rate of 33.38\% and 30.20\% for $N=1024$, and become invalid on $N=2048$. Transformer starts to lose some performance (99.94\%) on $N=1024$, and becomes not working when $N=2048$. Performer survives when $N\leq2048$, however, gets an error rate of 29.24\% for $N=4096$ and becomes as bad as random guessing for $N\geq8192$. Nystr\"omformer achieves 100\% accuracy for $N=4096$ and fails on $N\geq8192$. Instead, Paramixer reaches $100\%$ accuracy for all the tested lengths.

For the Temporal Order problem, the performance of all the models is 100\% or nearly 100\% when $N\leq256$, and starts to drop for $N\geq512$. When $N=2048$, the results of Performer (99.52\%) and Nystr\"omformer (100\%) are still solid while Transformer, Linformer, and Reformer have an error rate around 1.4\%. When increasing $N$ to 4,096 and 8,192, Linformer and Transformer fail in succession. When $N=16384$, Reformer does not work while Performer and Nystr\"omformer give a weaker performance with error rates, 7.6\% and 7.9\%, respectively. On the contrary, Paramixer achieves $100\%$ accuracy for $N\leq 8192$, and $98.84\%$ accuracy for $N=16384$.

In summary, Paramixer has better scalability than the attention neural networks based on scaled dot-products. When scaled to tens of thousands, Paramixer still gives very high prediction accuracy. The results suggest we can evaluate our model on more real-world tasks with very long sequences.

\setlength\tabcolsep{5pt}
\begin{table*}[t]
\centering
\caption{Classification accuracy by Paramixer and X-formers on the four LRA tasks. Methods that are absent in the corresponding paper are signed by a dash (``-''). For Paramixer, we present the mean ($\mu$) and standard deviation ($\sigma$) across multiple runs in the $\mu\pm\sigma$ format.}
\label{tab:LRAacc}
\begin{tabular}{lccccc}
\hline\hline
Model              & ListOps & Text & Image &  Pathfinder\\
                   & $N=2000$ & $N=4000$ & $N=1024$ & $N=1024$\\
\hline
Transformer \cite{tay2020long}    & 36.37 & 64.27 & 42.44 &  71.40\\
Transformer  \cite{zhu2021long}   & 37.13 & 65.35 & -     &  -    \\
Transformer \cite{nystromformer}    & 37.10 & 65.02 & 38.20 &  74.16\\
\hline
Sparse Transformer \cite{tay2020long}   & 17.07 & 63.58 & 44.24 & 71.71\\
\hline
Longformer \cite{tay2020long}     & 35.63 & 62.58 & 42.22 & 69.71\\
\hline
Linformer \cite{tay2020long}      & 37.70 & 53.94 & 38.56 & 76.34\\
Linformer \cite{zhu2021long}      & 37.38 & 56.12 & -     & -    \\
Linformer \cite{nystromformer}      & 37.25 & 55.91 & 37.84 & 67.60\\
\hline
Reformer \cite{tay2020long}       & 37.27 & 56.10 & 38.07 & 68.50\\
Reformer \cite{zhu2021long}       & 36.44 & 64.88 & -     & -    \\
Reformer \cite{nystromformer}       & 19.05 & 64.88 & 43.29 & 69.36\\
\hline
Performer \cite{tay2020long}      & 18.01 & 65.40 & 42.77 & \underline{77.05}\\
Performer \cite{zhu2021long}      & 32.78 & 65.21 & -     & -    \\
Performer \cite{nystromformer}      & 18.80 & 63.81 & 37.07 & 69.87\\
\hline
BigBird \cite{tay2020long}        & 36.06 & 64.02 & 40.83 & 74.87\\
\hline
Linear Transformer \cite{tay2020long}  & 16.13 & 65.90 & 42.34 & 75.30\\
\hline
Transformer-LS \cite{zhu2021long} & \underline{38.36} & 68.40 & -    &  -    \\
\hline
RFA-Gaussian \cite{peng2021random}   & 36.80 & 66.00 & -     & -    \\
\hline
Nystr\"omformer \cite{zhu2021long}  & 37.34 & 65.75 & -     & -    \\
Nystr\"omformer \cite{nystromformer}  & 37.15 & 65.52 & 41.58 & 70.94\\
\hline
Paramixer (CHORD) & \textbf{39.57}$\pm$0.32 & \underline{83.12}$\pm$0.33 &  \underline{45.01}$\pm$0.21 &  \textbf{80.49}$\pm$0.13\\
Paramixer (CDIL) & 37.78$\pm$0.28 & \textbf{83.32}$\pm$0.19 &  \textbf{46.58}$\pm$0.05 &  67.13$\pm$0.42\\
\hline
\hline
\end{tabular}
\vspace{3.4mm}
\end{table*}

\subsection{Long Range Arena Public Bechmark}

\setlength\tabcolsep{1.1pt}
\begin{figure}[t]
	\begin{center}
	\begin{tabular}{cccc}
		\includegraphics[width=0.12\textwidth]{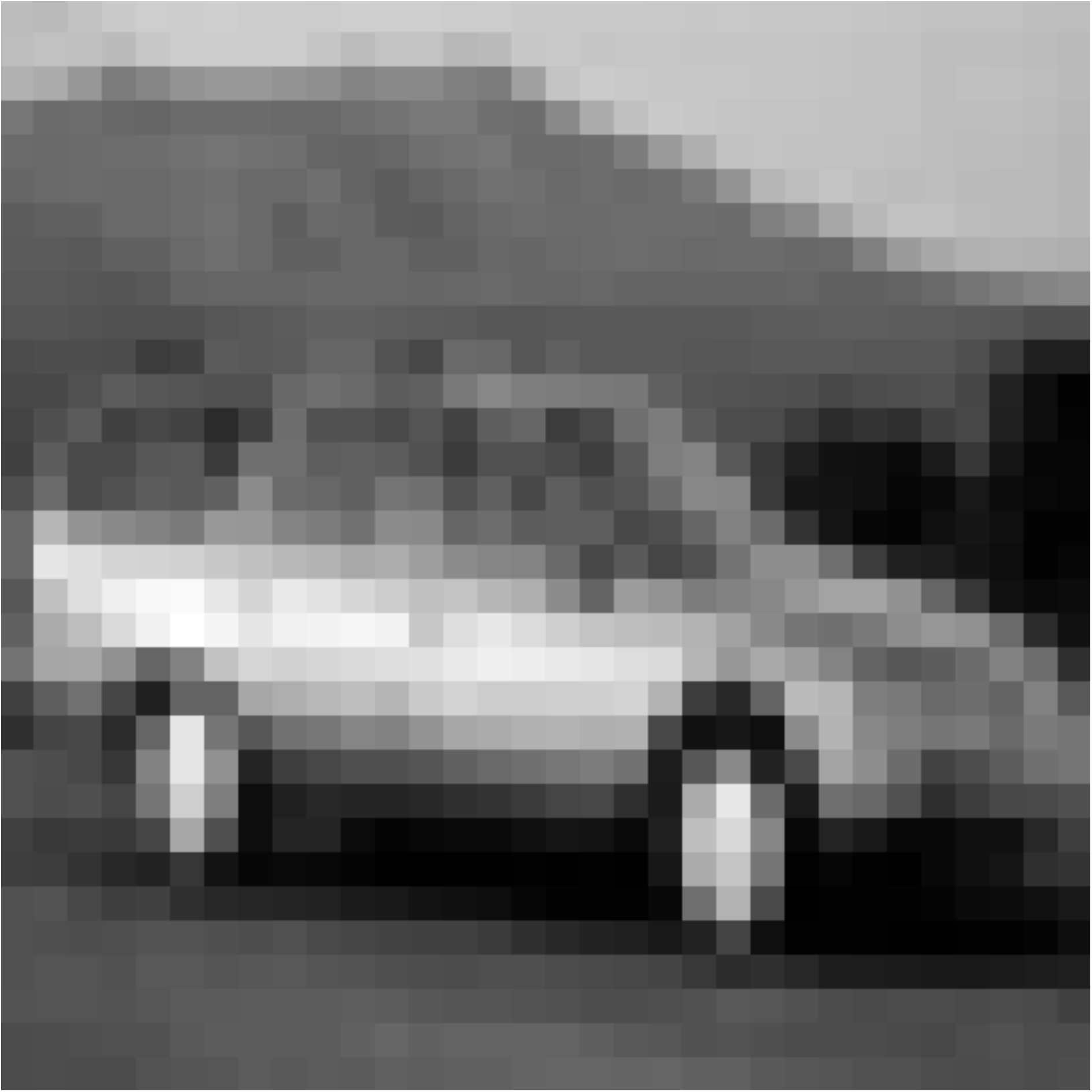} &
		\includegraphics[width=0.12\textwidth]{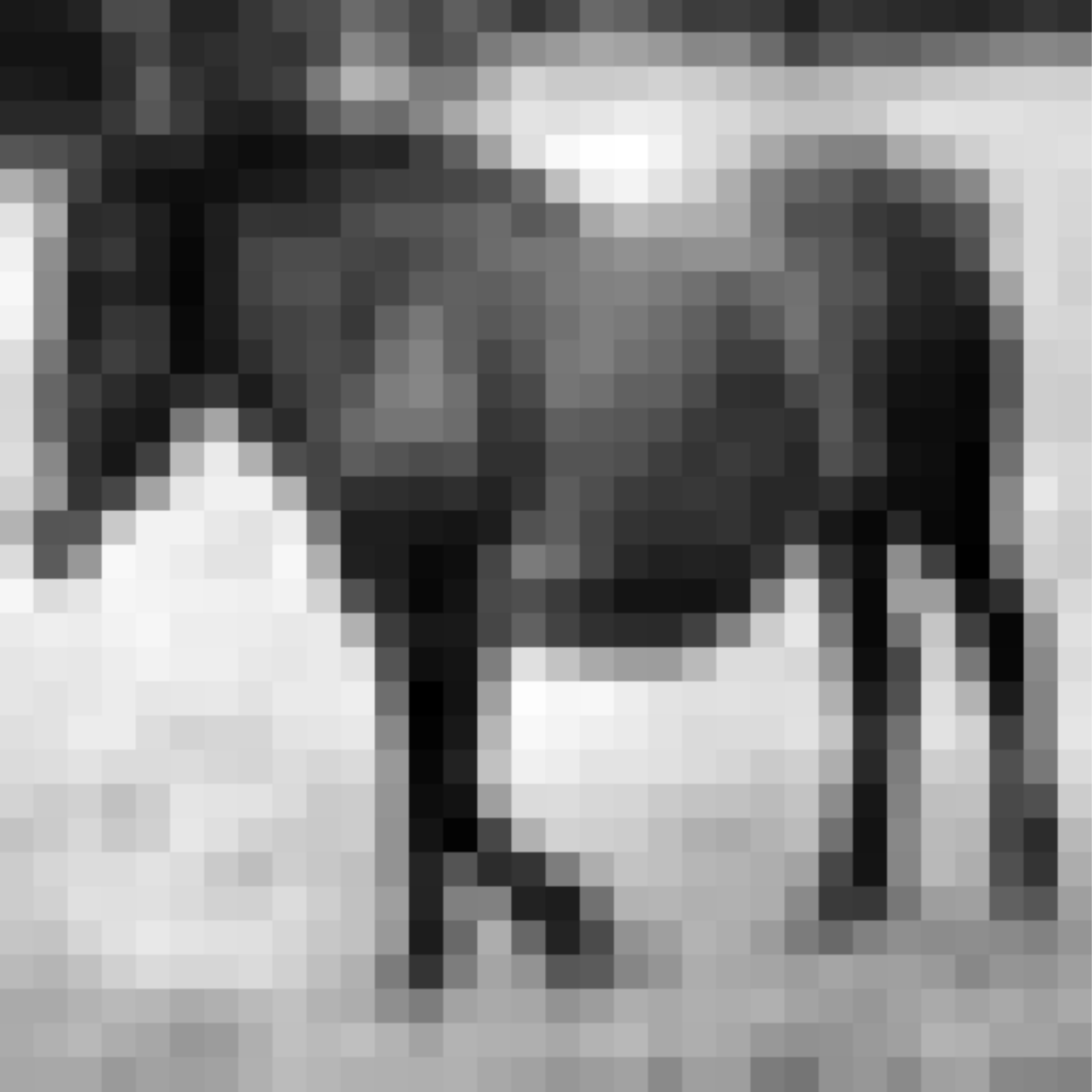} &
		\includegraphics[width=0.12\textwidth]{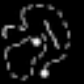} &
		\includegraphics[width=0.12\textwidth]{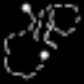}\\
		{\tiny {\bf Image, automobile}} &
		{\tiny {\bf Image, horse}}&
		{\tiny {\bf Pathfinder, Negative}} & 
		{\tiny {\bf Pathfinder, Positive}}\\
 	\end{tabular}
 	\vspace{2mm}
	\end{center}
	\caption{Example data from the Image Classification (left two) and Pathfinder tasks (right two).}
	\label{fig:lraex}
\end{figure}

Next, we evaluate the performance of Paramixer on Long Range Arena (LRA), a publicly available benchmark for modeling long sequential data \cite{tay2020long}. The details of the tasks are the following:
\begin{itemize}
    \item \emph{ListOps}. ListOps is a classification task designed for measuring the ability of models to parse hierarchically constructed data \cite{nangia2018listops}. Each sequence is composed of operators, digits, and left or right brackets. The brackets define lists of items. Each operator in a sequence takes the items in a list as input and returns a digit.
    \item \emph{Text Classification}. We use the IMDb Review dataset \cite{maas2011learning} which requires the model to classify each review as positive or negative. The task uses a character-level representation for each sequence, which makes the tasks more challenging than the word-level version. We truncated or padded every sequence to a fixed length ($N=4k$).
    \item \emph{Image Classification}. This task is to classify images into one of ten classes. Each image is flattened to form a sequence of length 1024. Unlike conventional computer vision, the task requires the predictors to treat the grayscale levels (0-255) as categorical values. That is, each image becomes a sequence of symbols with an alphabet size of 256. Two example matrices are shown in Figure \ref{fig:lraex}.
    \item \emph{Pathfinder}. This task is motivated by cognitive psychology \cite{linsley2018learning}, and constructed using synthetic images. Each image (size $32\times 32$) contains two highlighted endpoints and some path-like patterns. The models need to classify whether there is a path consisting of dashes between two highlighted points. Similar to the Image Classification task, the predictors must flatten the image to a sequence of symbols with length 1024. Two example matrices are shown in Figure \ref{fig:lraex}.
\end{itemize}

For a fair comparison, we followed the experiment settings in the original paper \cite{tay2020long} and evaluated multi-block ParamixerNet in the above tasks. We constructed Paramixer in variable blocks and used cross-validation to report the model with the best hyperparameters. We ran the model four times with a different random seed for each task. 

We compared Paramixer with many X-former architectures in prediction accuracies. If a method has different implementations, we quote all the alternatives and their results. The results show that Paramixer beats all the other self-attention-based transformers on all the tasks, having the best classification accuracy. Such stable cross-task wins suggest that forming an attention matrix with parameterizing mixing links works better than those based on scaled dot-products. 

Remarkably, Paramixer has achieved state-of-the-art performance on Text and Pathfinder tasks. For Text Classification, ParamixerNet achieves $83.32\%$, which is $14.92$ p.p. higher than the best Transformer variant, Transformer-LS ($68.4\%$). Our method also wins with the accuracy of $80.49\%$ on Pathfinder, where it gains about $5$ percentage points higher than the runner-up model. The compelling improvement brought by ParamixerNet is a probable cause of the high-rank nature of Natural Language \cite{yang2017breaking}.

For Paramixer, we reported two variants for this task, using the CHORD and CDIL protocols independently. As shown in Table \ref{tab:LRAacc}, CHORD wins on ListOps and Pathfinder, while CDIL gains the best results on Text and Image. However, CHORD still has a competitive accuracy on Text (83.12\%) and Image (45.01\%), which demonstrates that Paramixer has more stable performance using the CHORD protocol. Consequently, we used CHORD as a default protocol for the following experiments.

\subsection{Long Document Classification}

\setlength\tabcolsep{3pt}
\begin{table}[t]
\caption{Test accuracies on long document classification. }
\label{tab:othermats}
\begin{center}
\begin{tabular}{lccc}
\hline\hline\\[-3.5mm]
Model & $N = 16k$ & $N = 32k$ \\
\hline\\[-3mm]
 Transformer &        25.62 & 25.62  \\
 Linformer &          64.69  & 65.36  \\
 Performer &           25.62  & 25.20\\
 Reformer &            25.04  & 25.04\\
 Transformer-LS&          70.25 & 60.93 \\
 Nystr\"omformer &         71.70 & 67.69 \\
\hline\\[-3mm]
 Paramixer &             \bf{83.89} & \bf{84.55} \\
\hline
\hline
\end{tabular}
\vspace{2.8mm}
\end{center}
\label{docclas}
\end{table}

The goal of this subsection is to study the benefits of modeling long sequences in NLP tasks. The character-level text classification task from the LRA benchmark is not long enough to conclude this pattern, so we seek for Longer Document Data with tens of thousands of tokens.

Long-document-dataset is a publicly available dataset. It contains a collection of academic papers, which are parsed using the arXiv sanity preserver program and published on Github \cite{he2019long}. There are eleven diverse research areas a paper may belong to. Similar to the source article, we used only four classes of documents: cs.AI, cs.NE, math.AC, math.GR, having a final set of 11\,956 documents in total. We used 70\% of the data for training, 20\% for validation, and 10\% for the test. Unlike the original study \cite{he2019long}, we transformed each article into a sequence of characters and finally got a challenging task with an average training sequence length of 52\,112.

We truncate every sequence to a fixed length. Since NLP task is known to benefit from using longer sequence length, we tested Paramixer on sequence lengths of 16\,384 and 32\,768 to see if Paramixer can be in favor of using longer context. We studied other transformer-like variants as well.

As shown in Table \ref{docclas}, Paramixer outperforms the follow-up transformer-based competitors by 12.19 and 16.86 percentage points on two sequence lengths, respectively. Paramixer achieves higher accuracy when using 32k tokens, which signals a better generalization from using more extended context. Among all the competitors, only Linformer, Transformer-LS, and Nystr\"omformer can handle the sequences of this length to a certain degree. Nevertheless, the gap in performance between sequence lengths 32k and 16k is severe, suggesting that the low-rank factorization weakens the prediction power on high-rank NLP tasks.

\subsection{Genome Classification}

\setlength\tabcolsep{1.5pt}
\newcommand{\sqlenwidth}{7.5cm}
\begin{figure}[t]
    \centering
    \includegraphics[width=\sqlenwidth]{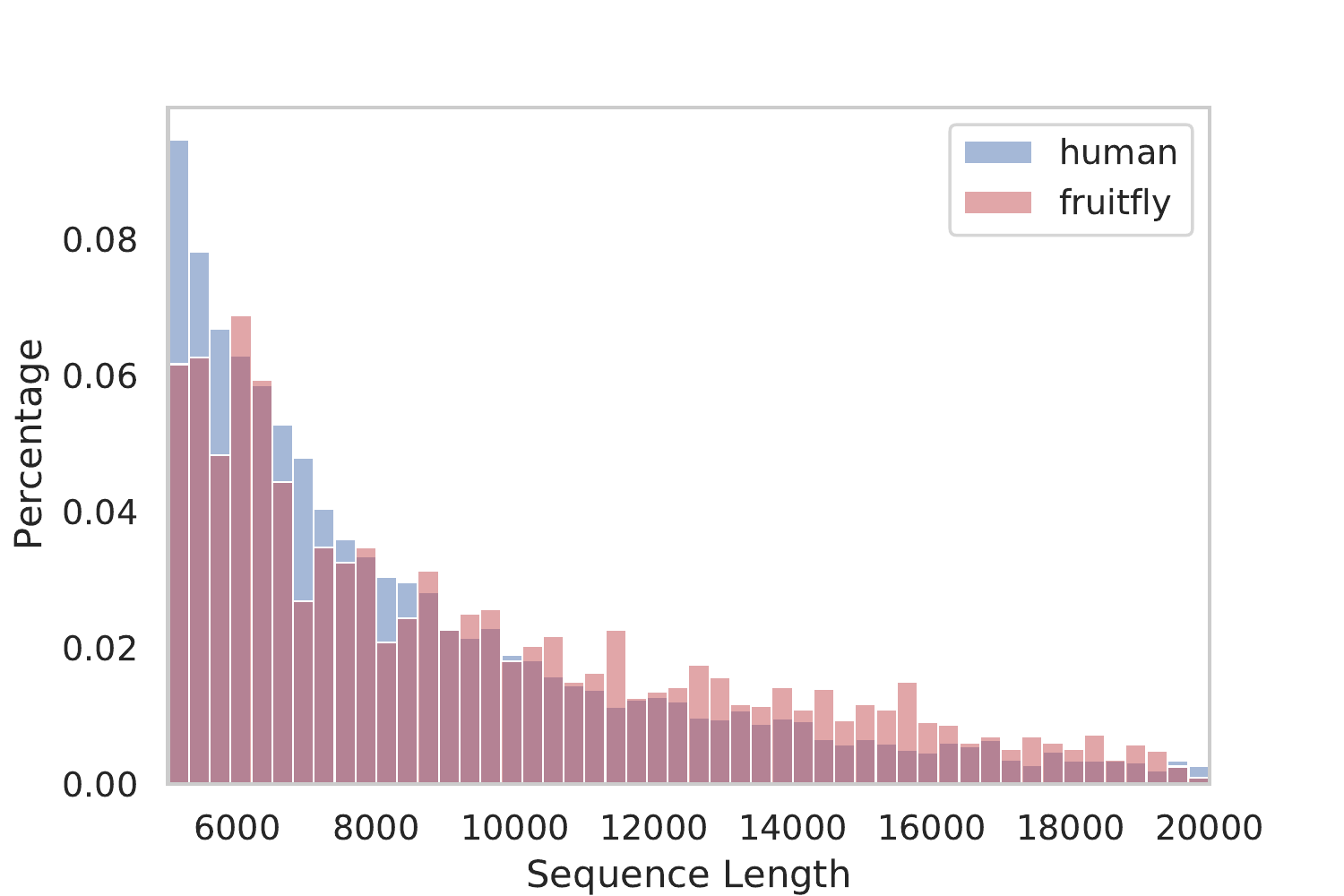}
	\includegraphics[width=\sqlenwidth]{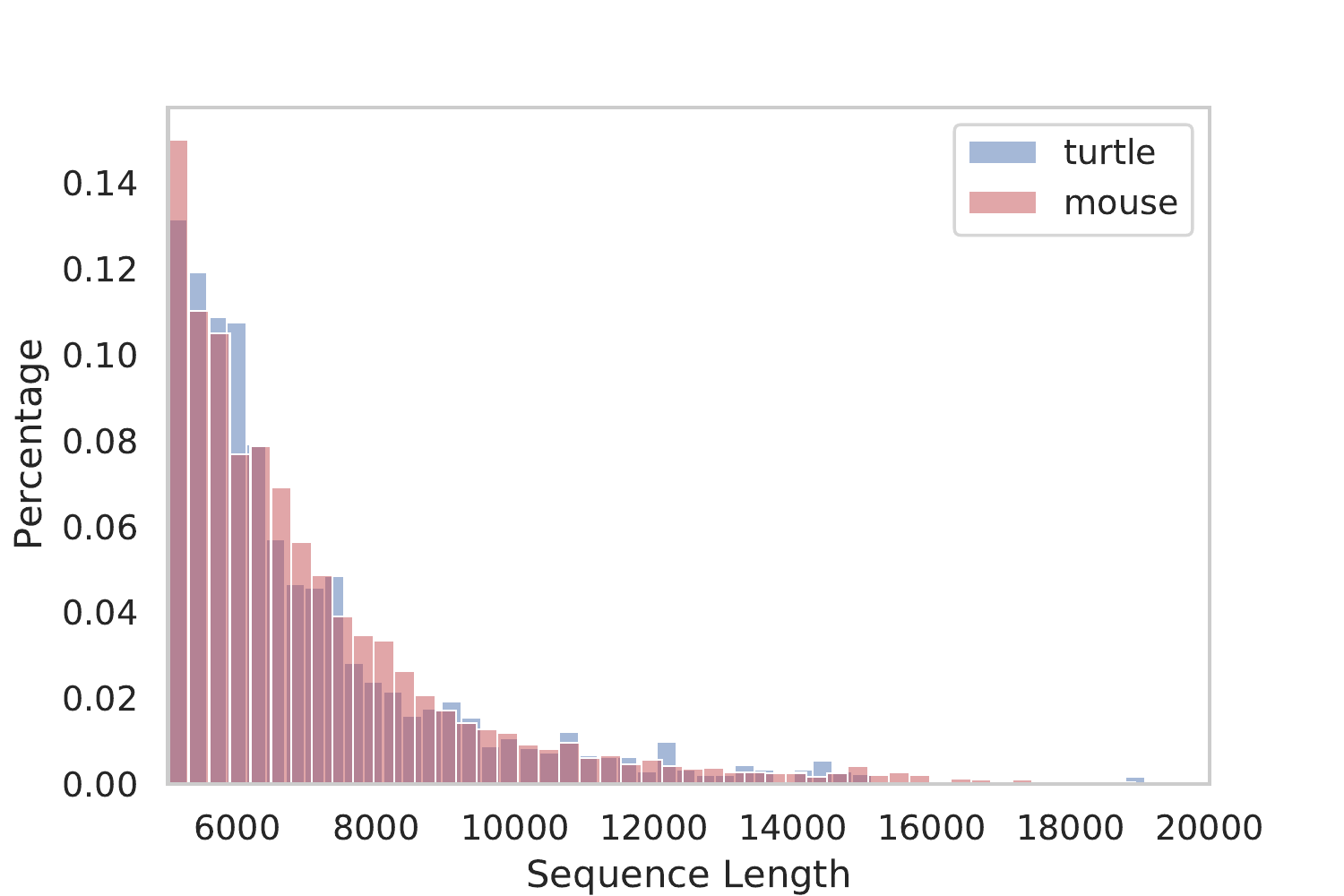}\\
	
    \caption{Sequence length distributions for HFDNA (top) and MTcDNA (bottom). X-axis is the range of genome sequence length, and y-axis is the percentage of total instance numbers for each bin.}
    \label{fig:lendist}
    \vspace{-1mm}
\end{figure}

 Inspired by a recent rise in using deep learning models in biological applications, such as Chromatin-profile prediction \cite{zaheer2020big} and genome analysis \cite{enformer}. However, directly processing genome and protein sequences with tens of thousands of positions are problematic. The standard preprocessing pipeline includes chunking a long sequence into many fragments, which are modeled independently. This approach is associated with inevitable information loss caused by the long-distant nature of interaction processes in the DNA sequences \cite{gasperini2020towards}. We tested our design on this type of data. By design, Paramixer can treat a full sequence as an input without the need for its preliminary segmentation, allowing it to capture highly non-local effects. 

We built two data sets for this task. The first group of DNA sequences were downloaded from NONCODEv6 \cite{noncodev6}\footnote{http://www.noncode.org/download.php}. We used human and fruitfly DNA sequences to construct a binary-classification task. We filtered out the sequences shorter than 5k and thus got 9\,536 human DNA sequences and 4401 fruitfly DNA sequences with mean sequence lengths 10\,586 and 9\,793, respectively. We name the data set HFDNA. As shown in Figure \ref{fig:lendist}, the two species in HFDNA have similar sequence length distributions, which means they can not be directly distinguished based on a sequence length threshold. We split the data set to 60\% training, 20\% validation, and 20\% test. Each sequence is either padded or truncated to a fixed length of 16\,384.

The other data set, namely MTcDNA (mouse and turtles' cDNA), was built following the same preprocessing/split strategy. The original cDNA sequences were downloaded from Ensembl genome browser\footnote{http://www.ensembl.org/info/data/ftp/index.html} \cite{howe2021ensembl, ensembl}. The task is to predict a binary class, given either mouse or turtles cDNA sequence. The mouse class presented in 12\,300 cDNA sequences with a mean sequence length of 7\,235, including two sub-species: \textit{Mus musculus} and \textit{Mus spretus}. The turtle class contains 4\,193 cDNA sequences of \textit{Chelonoidis abingdonii} and \textit{Gopherus agassizii} with a mean sequence length of 7\,072. The percentage histograms of sequence lengths are shown in Figure \ref{fig:lendist}. It is clearly seen that the two species have a big overlap in terms of sequence length, which makes it hard to discriminate between them only by length. 

We compared ParamixerNet with several transformer-based models on this task. Because both data sets are imbalanced, so we report ROC AUC values for this experiment. As shown in Table \ref{gencls},  Paramixer achieves 100\%  ROC AUC on HFDNA, outperforming the runner-up result from Transformer (94.32\%). Notably, our design holds the best score at 84.86\% ROC AUC on MTcDNA, which is higher than Transformer by 5.48 percentage points. The experimental results show that Paramixer can successfully handle long DNA sequences and has the potential to assist in the biological applications that require modeling long-distant gene interactions.

\setlength\tabcolsep{3pt}
\begin{table}[t]
\caption{The performance of Paramixer in comparison to Xformers on the Genomic classification task. The metric is Area Under the Receiver Operating Characteristic Curve (ROC AUC) $\times 100\%$}
\label{tab:othermats}
\begin{center}
\begin{tabular}{lcc}
\hline\hline\\[-3.5mm]
Model & HFDNA & MTcDNA\\
\hline\\[-3mm]
 Transformer &        94.32 & 79.38 \\
 Linformer &          91.65 & 77.06 \\
 Performer &           93.46 & 78.92\\
 Reformer &            92.32 & 74.94 \\
 Transformer-LS&          93.19 & 75.81\\
 Nystr\"omformer &         92.26 & 71.98\\
\hline\\[-3mm]
     Paramixer &             \bf{100.00} & \bf{84.86} \\
\hline
\hline
\end{tabular}
\vspace{-2mm}
\end{center}
\label{gencls}
\end{table}

\section{Conclusion}
\label{sec:conclusion}
We have proposed a scalable and effective building block called Paramixer for attention neural networks. Our method replaced the dot-products and softmax with parameterization of the mixing links in full-rank sparse factors of the attention matrix and thus got rid of the low-rank bottleneck in most existing attention models. Our method has complexity as low as $O(N\log N)$ and can efficiently deal with sequences up to tens of thousands. Besides scalability, Paramixer has also demonstrated strong performance in terms of accuracy. Neural networks, by stacking the proposed Paramixer blocks, have defeated Transformer and many of its variants in a variety of tasks, including synthetic data inference, the public Long Rang Arena benchmark, and classification of very long text documents and genome sequences.

In the future, we could study other applications beyond classification, for example, gene expression prediction from sequence and pretraining with unsupervised data. The basic Paramixer block could be extended using other existing attention techniques such as multiple heads and relative positional encoding. Later, in addition to the CHORD and CDIL protocols, we could consider the other predefined protocols or even adaptively learned protocols for the sparse structure.

\bibliographystyle{unsrt}  
\bibliography{main}

\clearpage
\begin{center}
\textbf{\large Appendix. Sparse Factorization of Large Square Matrices}
\end{center}
\setcounter{equation}{0}
\setcounter{figure}{0}
\setcounter{table}{0}
\setcounter{page}{1}
\setcounter{section}{0}
\makeatletter
\renewcommand{\theequation}{A\arabic{equation}}
\renewcommand{\thefigure}{A\arabic{figure}}
\renewcommand{\thetable}{A\arabic{table}}

\section{Synthetic Data Experiments}
For both problems in the scalability test, we generated sequences using the setup described in the main paper. We ran the experiments on sequences with variable lengths: from 128 to 32k. The longer sequences, the more complex the retrieval process. There is a slight difference in the pre-processing part. For the Adding problem, the input data was only two-dimensional. To avoid using such a low-dimensional embedding space, we augmented the dimensionality with an additional linear layer to assure sufficient freedom for dot-product attention architectures. The training configuration and hyperparameters are the same for both the Adding problem and the Temporal Order problem. Their summary is in Table \ref{tab:configs}

\section{Long Range Arena}

The data set for the LRA benchmark is publicly available. The information about data and the download link can be found in the official GitHub repository: \url{https://github.com/google-research/long-range-arena}.

\begin{itemize}
    \item \textbf{ListOps} The raw data for this problem is organized as three separate files \texttt{basic\_train.tsv}, \texttt{basic\_test.tsv}, \texttt{basic\_val.tsv} for training, testing, and validation data, respectively. The split is fixed. In addition to the tokens described in the main paper, each sequence has "(" and ")" symbols, which should be removed. To equalize the lengths of the sequences, we used the built-in PyTorch padding functional. After the sequences are prepared, the embedding layer processes each unique value, thus mapping elements to the embedding space. The rest of the training process is straightforward. 
    \item \textbf{Text Classification} We downloaded IMDB data set using the \texttt{tensorflow-dataset} package, and got 25000 instances for training and another 25000 for testing. We went through the whole corpus and extracted the character vocabulary. Then we mapped each sequence to a vector of indices using this vocabulary. Finally, we truncated or padded each sequence to a fixed length of 4096. For every review, we add [”CLS”] token to each sequence and use the embedding of ["CLS"] token for final classification. We used three blocks Paramixer for this task. 
    \item \textbf{Image Classification} CIFAR10 is a well-known dataset, which can be downloaded from the \texttt{torchvision} package. The train/test splitting is fixed. To make images grayscaled, we used standard transformation \texttt{transforms—grayscale} from the same package. An image is flattened to a sequence of length 1024. Then each element is mapped to a dictionary of size 256 (all possible intensity values) and given to the embedding layer.  
    \item \textbf{Pathfinder} The problem data consists of two types of files: images and metafiles. Metafiles store information about all the images and their corresponding labels (positive or negative). There are three classes of images: \texttt{curv\_baseline} (easy), \texttt{curv\_contour\_length\_9} (medium), \texttt{curv\_contour\_length\_14} (hard). An image class corresponds to the distance between its endpoints (curve length), thus positively correlates with the difficulty level. The exact data split is not provided. To separate the data into three parts, we iterated over all metafiles from the catalogs and constructed the training/val/test (90\%/5\%/5\%) sets such that all three types of images are present equally. The rest of the processing is similar to the Image Classification task.
\end{itemize}

\section{Long Document Classification}
The task is a four-class classification problem. The class of a paper is defined by its arxiv categorization, namely, cs.AI, cs.NE, math.AC, and math.GR. Each class in the data set is almost equally presented, with a slight class imbalance: 2995, 3012, 2885, and 3065 documents, respectively. To transform the raw articles into sequences, we first went through the whole corpus and extracted the character vocabulary. Then we mapped each character sequence to a vector of indices using this vocabulary. We fine-tuned Paramixer and X-formers to get the best results. The hyperparameters of Paramixer were selected using a similar process. Final configurations are shown in Table \ref{tab:configs}. For every document, we add [”CLS”] token to each sequence and use the result embedding of ["CLS"] token for the final classification.

\section{Genome Classification}
When building MTcDNA we downloaded cDNA sequences of Chelonoidis abingdonii and Gopherus agassizii, and merged them as a turtle data set. Following the same strategy, we built the mouse data set using Mus musculus and Mus spretus. More details can be found in the main paper. For the HFDNA classification task, one Paramixer block is enough to get 100\% accuracy. However, ParamixerNn with two blocks result in the best test accuracy for MTcDNA. The selected hyperparameters are listed in Table \ref{tab:configs}. 

\begin{table*}[t]
\centering
\caption{Hyperparameters details for every task. $N$, $B$, $V$, $E$, $H$, lr refer to max sequence length, batch size, vocabulary size, embedding size, hidden states size, and learning rate, respectively. The vocabulary size includes padding index and ["CLS"].}
\label{tab:configs}
\begin{tabular}{lccccccccccc}
\hline\hline
Task     & $N$ & Protocol & n\_links & lr & $B$ & $V$ & $E$ & $H$ & pos\_embed & Pooling Type \\
\hline
Adding         & 32768 & CHORD & 15 & 0.001 & 40 & - & 32 & 32 & True & FLAT\\
Temporal Order & 16384 & CHORD & 14 & 0.001 & 40 & 6 & 32 & 32 & True & FLAT\\
\hline
ListOps & 2000 & CHORD & 12 & 0.001 & 48 & 16 & 32 & 32 & True & FLAT\\
CIFAR10 & 1024 & CDIL & 3 & 0.001 & 64 & 256 & 32 & 32 & True & FLAT\\ 
Text & 4096 & CDIL & 9 & 0.0001 & 32 & 97 & 32 & 128 & False & CLS\\
Pathfinder & 1024 & CHORD & 11 & 0.001 & 64 & 256 & 32 & 32 & True & FLAT\\ 
\hline
Long Document & 16384 & CHORD & 15 & 0.0001 & 16 & 4290 & 100 & 128 & False & CLS\\
Long Document & 32768 & CHORD & 16 & 0.0001 & 16 & 4290 & 100 & 128 & False & CLS\\
\hline
Genome Classification & 16384 & CHORD & 15 & 0.0001 & 16 & 5 & 32 & 128 & True & FLAT\\
\hline
\hline
\end{tabular}
\end{table*}

\section{Proof of Proposition 3.1 in the main paper}
\begin{definition}
An $N\times N$ circulant matrix $C$ takes the form
\begin{align*}
    C = \left[
    \begin{array}{ccccc}
         c_0 & c_{N-1} & \cdots & c_2 & c_1 \\
         c_1 & c_0 & c_{N-1} & \cdots & c_2 \\
         \vdots & c_1 & c_0 & \ddots & \vdots \\
         c_{N-2} & \cdots & \ddots & \ddots & c_{N-1} \\
         c_{N-1} & c_{N-2} & \ddots & c_1 & c_0
    \end{array}
    \right]
\end{align*}
\end{definition}

\begin{definition}
The polynomial 
\begin{align*}
f(x)=c_0+c_1x+\dots+c_{N-1}x^{N-1}    
\end{align*}
is called the associated polynomial of circulant matrix $C$.
\end{definition}

We have the following theorem in the literature \cite{circulantrank}:
\begin{theorem}
The rank of a circulant matrix $C$ is equal to $N-d$, where $d$ is the degree of the polynomial $\text{GCD}(f(x), x^{N-1})$.
\end{theorem}

Now we can prove Preposition 3.1 for the CHORD protocol. The proof for CDIL follows similarly.
\begin{proof}
The associated polynomial of $W^{(m)}$ is
\begin{align*}
    f(x) = \sum_{k=0}^{\log_2N-1}x^k
\end{align*}
Because $\text{GCD}(f(x), x^N-1)=1=x^0$, the rank of $W^{(m)}$ is $N$.
\end{proof}

\end{document}